\documentclass[letterpaper, 10 pt, conference]{ieeeconf}  % Comment this line out if you need a4paper

\IEEEoverridecommandlockouts                              % This command is only needed if 
\usepackage{graphicx}
\usepackage{amsmath,amssymb}
\usepackage{caption}
\usepackage{subcaption}
\usepackage{float}
\usepackage{array}
\usepackage{multirow}

\newcolumntype{P}[1]{>{\centering\arraybackslash}p{#1}}
\usepackage{booktabs}
\newcommand{\ra}[1]{\renewcommand{\arraystretch}{#1}}

\title{\LARGE \bf
Visual Stability Prediction and Its Application to Manipulation
}

\author{Wenbin Li$^{1}$, Ale\v{s} Leonardis$^{2}$,  Mario Fritz$^{1}$% <-this % stops a space
\thanks{$^{1}$Max Planck Institute for Informatics, Saarland Informatics Campus, Germany
{\tt\small \{wenbinli,mfritz\}@mpi-inf.mpg.de}}%
\thanks{$^{2}$School of Computer Science, University of Birmingham, United Kingdom
{\tt\small a.leonardis@cs.bham.ac.uk}}
}

\begin{document}
\bstctlcite{IEEEexample:BSTcontrol}

\maketitle
\thispagestyle{empty}
\pagestyle{empty}

%%%%%%%%%%%%%%%%%%%%%%%%%%%%%%%%%%%%%%%%%%%%%%%%%%%%%%%%%%%%%%%%%%%%%%%%%%%%%%%%
\begin{abstract}

Understanding physical phenomena is a key competence that enables humans and animals to act and interact under uncertain perception in previously unseen environments containing novel objects and their configurations. Developmental psychology has shown that such skills are acquired by infants from observations at a very early stage.

In this paper, we contrast a more traditional approach of taking a model-based route with explicit 3D representations and physical simulation by an {\em end-to-end} approach that directly predicts stability from appearance. We ask the question if and to what extent and quality such a skill can directly be acquired in a data-driven way---bypassing the need for an explicit simulation at run-time.

We present a learning-based approach based on simulated data that predicts stability of towers comprised of wooden blocks under different conditions and quantities related to the potential fall of the towers. We first evaluate the approach on synthetic data and compared the results to human judgments on the same stimuli. Further, we extend this approach to reason about future states of such towers that in turn enables successful stacking.

\end{abstract}

\section{Introduction}
Scene understanding requires, among others, understanding of relations between and among the objects. Many of these relations are governed by the Newtonian laws and thereby rule out unlikely or even implausible configurations for the observer. They are part of ``dark matter'' \cite{Xie_2013_ICCV} in our everyday visual data which helps us {\em interpret} the configurations of objects correctly and accurately. 
Although objects simply obey these elementary laws of Newtonian mechanics, which can very well be captured in simulators, uncertainty in perception makes exploiting these relations challenging in artificial systems.

In contrast, humans understand such physical relations naturally, which e.g., enables them to manipulate and interact with objects in unseen conditions with ease. We build on a rich set of prior experiences that allow us to employ a type of commonsense understanding that, most likely, does not involve symbolic representations of 3D geometry that is processed by a physics simulation engine. We rather build on what has been coined as ``na\"{i}ve physics'' \cite{Smith1994} or ``intuitive physics'' \cite{mccloskey1983intuitive}, 
which is a good enough proxy to make us operate successfully in the real-world.

\setlength{\belowcaptionskip}{-10pt}
\begin{figure}
\centering
\includegraphics[width=0.95\linewidth]{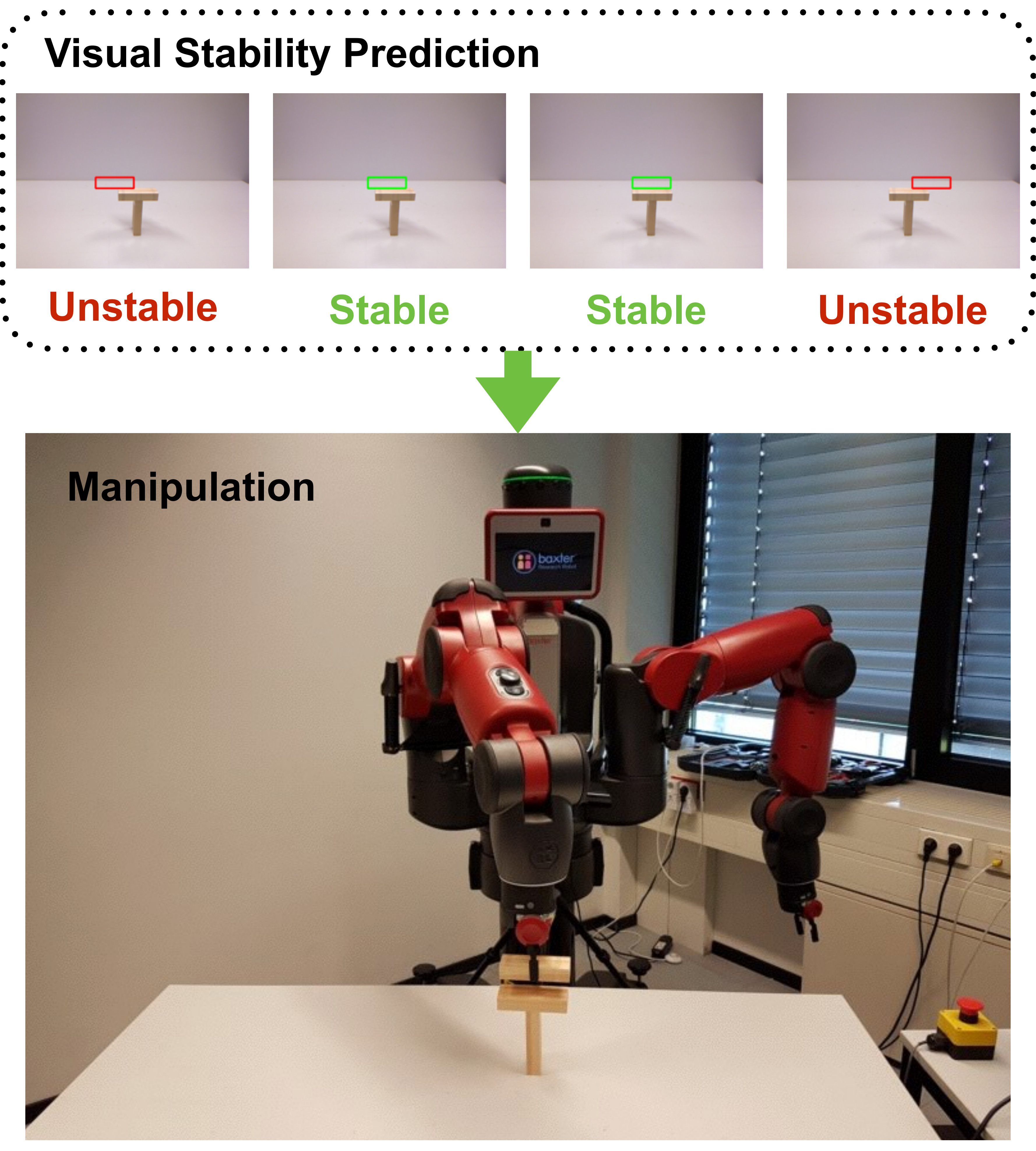}
\caption{Given a wood block structure, our visual stability classifier predicts the  stability for future placements, the robot then stacks a block among the predicted stable placements.}
\label{fig:teaser}
\end{figure}
\setlength{\belowcaptionskip}{-2pt}

It has not yet been shown how to equip machines with a similar set of physics commonsense -- and thereby bypassing a model--based representation and a physical simulation. In fact, it has been argued that such an approach is unlikely due to e.g., the complexity of the problem \cite{battaglia2013simulation}. Only recently, several works have revived this idea and reattempted a fully data drive approach to capturing the essence of physical events via machine learning methods \cite{mottaghi2015newtonian,wu2015galileo,fragkiadaki2015learning}.

In contrast, studies in developmental psychology \cite{baillargeon1994infants} have shown that infants acquire knowledge of physical events by observation at a very early age, for example, support, how an object can stably hold another object; collision, how a moving object interact with another object. According to their research, the infant with some innate basic core knowledge \cite{baillargeon2008innate} gradually builds its internal model of the physical event by observing its various outcomes. Amazingly, such basic knowledge of physical event, for example the understanding of support phenomenon can make its way into relatively complex operations to construct structures. Such structures are generated by stacking up an element or removing one while retaining the structure's stability primarily relying on effective knowledge of support events in such toy constructions. In our work, we focus on exactly this support event and construct a model for machines to predict object stability.

We revisit the classic setup of Tenenbaum and colleagues \cite{battaglia2013simulation} and explore to which extend machines can predict physical stability events directly from appearance cues. 
We approach this problem by synthetically generating a large set of wood block towers under a range of conditions, including varying number of blocks, varying block sizes, planar vs.\ multi-layered configurations. We run those configurations through a simulator ({\em only at training time!}) in order to generate labels whether the tower would fall or not. We show for the first time that aforementioned stability test can be learned and predicted in a purely data driven
way---bypassing traditional model-based simulation approaches. Further, we accompany our experimental study with human judgments on the same stimuli. 

We also apply the approach to guide the robot to stack blocks based on the stability prediction. To circumvent the domain shift between the synthesized images and the real world scene images, we extract the foreground masks for both synthesized and captured images. Given a real world block structure, the robot uses the model trained on the synthesized data to predict the stability outcome across possible candidate placements, and performs stacking on the feasible locations afterwards. We evaluate both the prediction and manipulation performance on the very task.
\section{Related Work}As humans, we possess the ability to judge from vision alone if an object is physically stable or not and predict the objects' physical behaviors. Yet it is unclear: (1) How do we make such decisions and (2) how do we acquire this capability? Research in development psychology \cite{baillargeon1994infants,baillargeon1995model,baillargeon2002acquisition} suggests that infants acquire the knowledge of physical events at very young age by observing those events, including support events and others. This partly answers Question 2, however, there seems to be no consensus on the internal mechanisms for interpreting external physical events to address Question 1. Battaglia et {\em al.}, \cite{battaglia2013simulation} proposed an intuitive physics simulation engine for such a mechanism and found that it resembles behavior patterns of human subjects on several psychological tasks. 
Historically, intuitive physics is connected to the cases where people often hold erroneous physical intuitions \cite{mccloskey1983intuitive}, such as they tend to expect an object dropped from a moving subject to fall vertically straight down. It is rather counter-intuitive how the proposed simulation engine in \cite{battaglia2013simulation} can explain such erroneous intuitions. 

While it is probably illusive to fully reveal human's inner mechanisms for physical modeling and inference, it is feasible to build up models based on observation, in particular the visual information. In fact, looking back to the history, physical laws were discovered through the  observation of physical events \cite{macdougal2012galileo}. Our work is in this direction. By observing a large number of support event instances in simulation, we want to gain deeper insights into the prediction paradigm.  

In our work, we use a game engine to render scene images and a built-in physics simulator to simulate the scenes' stability behavior. The data generation procedure is based on the platform used in \cite{battaglia2013simulation}, however as discussed before, their work hypothesized a simulation engine as an internal mechanism for human to understand the physics in the external world while we are interested in finding an image-based model to directly predict the physical behavior from visual channel. Learning from synthetic data has a long tradition in computer vision and has recently gained increasing interest \cite{li12eccv,kostas14cvpr,peng2015learning,rematas16cvpr} due to data hungry deep-learning approaches.

Understanding physical events also plays an important role in scene understanding in computer vision. By including additional clues from physical constraints into the inference mechanism, mostly from the support event, it has further improved results in segmentation of surfaces \cite{gupta2010blocks}, scenes \cite{silberman2012indoor} from image data, and object segmentation in 3D point cloud data \cite{zheng2013beyond}.

Only very recently, learning physical concepts from data has been attempted.
Mottaghi et {\em al.} \cite{mottaghi2015newtonian} aim at understanding dynamic events governed by laws of Newtonian physics, but use proto-typical motion scenarios as exemplars.
In \cite{fragkiadaki2015learning}, they analyze billiard table scenarios and aim at learning the dynamics from observation. While learning is largely data-driven, the object notion is predefined as location of the balls are provided to the system.
\cite{wu2015galileo} aims to understand physical properties of objects. They again rely on an explicit physical simulation. In contrast, we only use simulation at training time and predict for the first time visual stability directly from visual inputs of scenes containing various towers with a large number of degrees of freedom.

In \cite{fergus16blocsarxiv}, the authors present their work similar to our setting. Yet the focus of their work is different from ours, namely predicting outcome and falling trajectories for simple 4 block scenes, whereas  we significantly vary the scene parameters, investigating if and how the prediction performance from image trained model changes according to such changes, and further we examine how the human's prediction adapt to the variation in the generated scenes and compare it to our model.

To shed more light on the capabilities and limitations of our model, we explore how it can be used in a real robotic application, i.e., stacking a wood block given a block structure. In the past, we have seen researchers perform tasks with wood blocks, like playing Jenga from different perspectives. \cite{kroger2006demonstration} demonstrated multi-sensor integration by using a marker-based system with multiple cameras and sensors: a random block is first chosen in the tower, then the robot arm will try to pull the very block, if the force sensor detects large counter force or the CCD cameras detect large motion of tower, then the robot will stop pulling and try other block. \cite{wang2009robot} improved on \cite{kroger2006demonstration} by further incorporating a physics engine to initialize the candidates for pulling test. A different line of research is \cite{kimura2010force} where physical force is explicitly formulated with respect to the tower structure for planning. In our work, we do not do explicit formation of contact force as in \cite{kimura2010force}, nor do we perform trials on-site for evaluating the robot's operation. We only use physics engine to acquire synthesized data for training the visual-physics model. At test time, the planning system for our robot mainly exploits the knowledge encoded in the visual-physics model to evaluate the feasibility of individual candidates and performs operations accordingly. 
\section{Visual Stability Prediction}

\begin{figure}
\centering
\includegraphics[width=1.\linewidth]{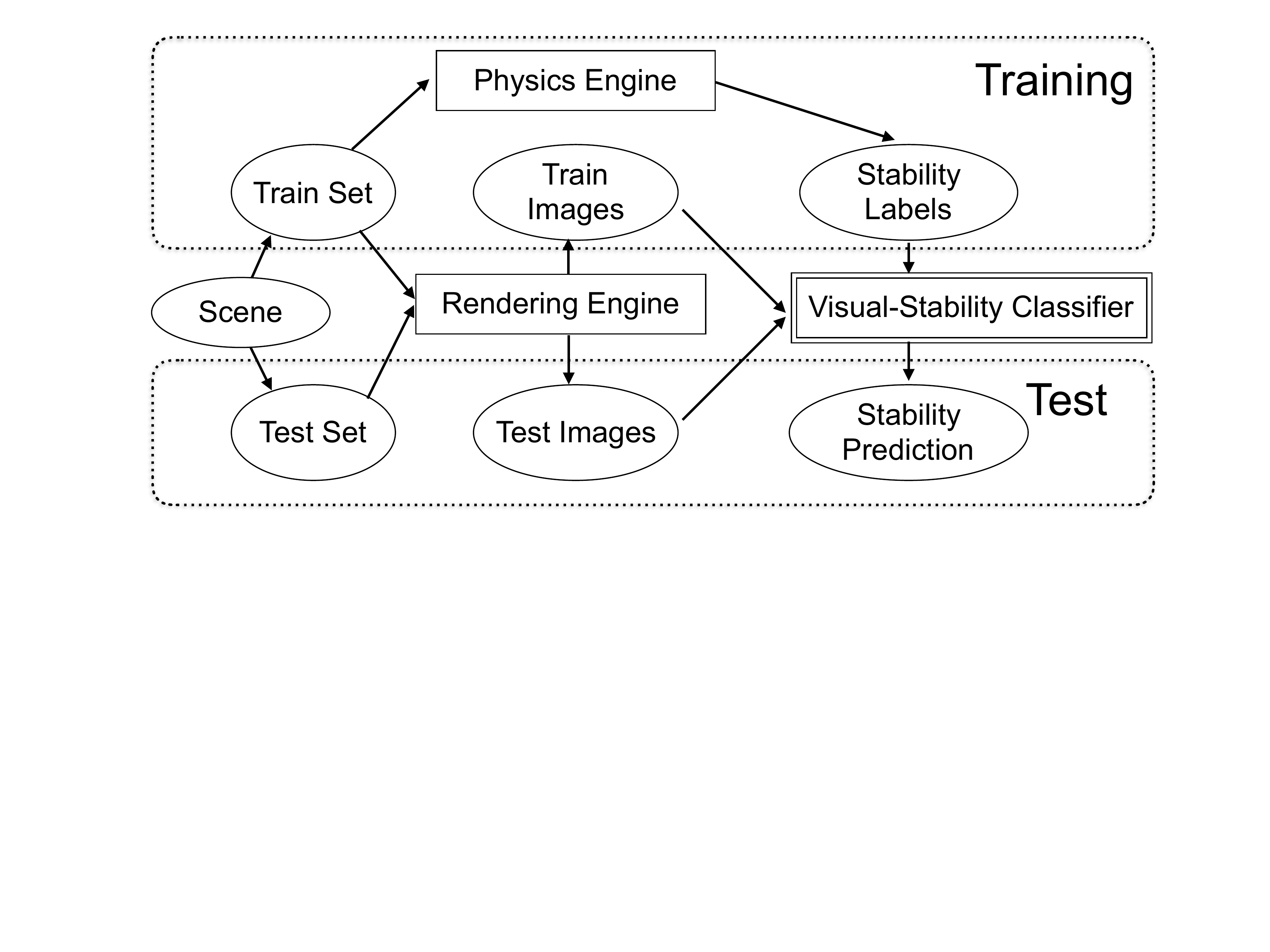}
\caption{An overview of our approach for learning visual stability. Note that physics engine is only used during training time to get the ground truth to train the deep neural network while at test time, only rendered scene images are given to the learned model to predict the physical stability of the scenes.}
\label{fig:overview_recog}
\end{figure}

In order to tackle a visual stability test, we require a data generation process that allows us to control various degrees of freedom induced by the problem as well as generation of large quantities of data in a repeatable setup. Therefore, we follow the seminal work on this topic \cite{battaglia2013simulation} and use a simulator to setup and predict physical outcomes of wood block towers. Afterwards, we describe the method that we investigate for visual stability prediction. We employ state-of-the-art deep learning techniques, which are the de-facto standard in today's recognition systems. Lastly, we describe the setup of the human study that we conduct to complement the machine predictions with a human reference. An overview of our approach is shown in Figure~\ref{fig:overview_recog}.

\subsection{Synthetic Data}
Based on the scene simulation framework used in \cite{hamrick2011internal,battaglia2013simulation}, we also generate synthetic data with rectangular cuboid blocks as basic elements. The number of blocks, blocks' size and stacking depth are varied in different scenes, to which we will refer as {\it scene parameters}. 
\paragraph{Numbers of Blocks}
We expect that varying the size of the towers will influence the difficulty and challenge the competence of ``eye-balling'' the stability of a tower in humans and machine. While evidently the appearance becomes more complex with the increasing number of blocks, the number of contact surfaces and interactions equally make the problem richer. Therefore, we include scenes with four different number of blocks, i.e., 4 blocks, 6 blocks, 10 blocks and 14 blocks as $\{{\text 4B}, {\text 6B}, {\text 10B}, {\text 14B}\}$.

\paragraph{Stacking Depth}
As we focus our investigations on judging stability from a monocular input, we vary the depth of the tower from a one layer setting which we call ${\text 2D}$ to a multi-layer setting which we call ${\text 3D}$.
The first one only allows a single block along the image plane at all height levels while the other does not enforce such constraint and can expand in the image plane. Visually, the former results in a single-layer stacking similar to Tetris while the latter ends in a multiple-layer structure as shown in Table~\ref{tab:scene_param}. The latter most likely requires the observer to pick up on more subtle visual cues, as many of its layers are heavily occluded. 

\paragraph{Block Size}
We include two groups of block size settings. In the first one, the towers are constructed of blocks that have all the same size of $1 \times 1 \times 3$ as in the \cite{battaglia2013simulation}. The second one introduces varying block sizes where two of the three dimensions are randomly scaled with respect to a truncated Normal distribution $N(1,\sigma^2)$ around $[1 - \delta,1 + \delta]$,  $\sigma$ and $\delta$ are small values. These two settings are referred to as $\{{\text Uni}, {\text NonUni}\}$. The setting with non-uniform blocks introduces small visual cues where stability hinges on small gaps between differently sized blocks that are challenging even for human observers.

\paragraph{Scenes}
Combining these three scene parameters, we define $16$ different scene groups. For example, group 10B-2D-Uni is for scenes stacked with 10 Blocks of same size, stacked within a single layer. For each group, $1000$ candidate scenes are generated where each scene is constructed with non-overlapping geometrical constraint in a bottom-up manner. There are $16 {\text K}$ scenes in total. For prediction experiments, half of the images in each group are for training and the other half for test, the split is fixed across the experiments.

\paragraph{Rendering}
While we keep the rendering basic, we like to point out that we deliberately decided against colored bricks as in \cite{battaglia2013simulation} in order to challenge perception and make identifying brick outlines and configurations more challenging. The lighting is fixed across scenes and the camera is automatically adjusted so that the whole tower is centered in the captured image. Images are rendered at resolution of $800 \times 800$ in color.

\paragraph{Physics Engine}
We use Bullet \cite{coumans2010bullet} in Panda3D \cite{goslin2004panda3d} to perform physics-based simulation for $2\text{s}$ at $1000{\text Hz}$ for each scene. Surface friction and gravity are enabled in the simulation. The system records the configuration of a scene of $N$ blocks at time $t$ as $(p_1,p_2,...,p_N)_t$, where $p_i$ is the location for block $i$. The stability is then automatically decided as a Boolean variable: 
\begin{gather*}
\begin{aligned}
S = \bigvee\limits_{i=1}^N (\Delta((p_i)_{t=T} - (p_i)_{t=0}) > \tau)
\end{aligned}
\end{gather*}
where $T$ is the end time of simulation, $\delta$ measures the displacement for the blocks between the starting point and end time, $\tau$ is the displacement threshold, $\bigvee$ denotes the logical ${\text{Or}}$ operator, that is to say it counts as unstable $S={\text{True}}$ if any block in the scene moved in simulation, otherwise as stable $S={\text {False}}$.

\begin{table*}[]
\begin{tabular}{cccc|cc|cc}
\hline\hline
\multicolumn{4}{c}{Block Numbers} & \multicolumn{2}{c}{Stacking Depth} &\multicolumn{2}{c}{Block Size}\\
\hline
& & & & & & & \\
\begin{subfigure}{0.101\textwidth}\centering\includegraphics[width=0.96\columnwidth]{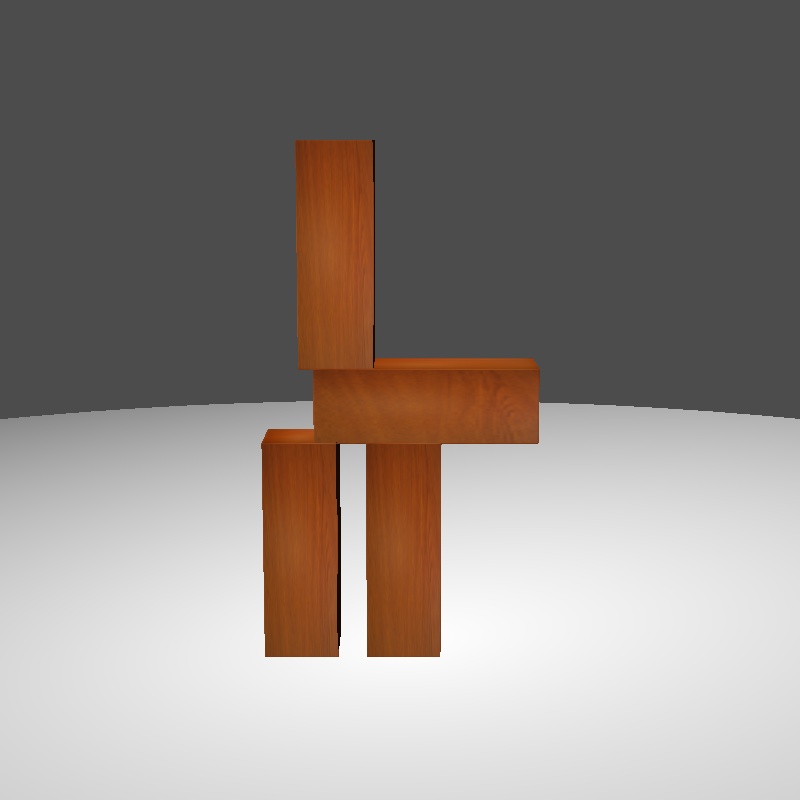}\caption{4 Blocks}\end{subfigure}&
\begin{subfigure}{0.101\textwidth}\centering\includegraphics[width=0.96\columnwidth]{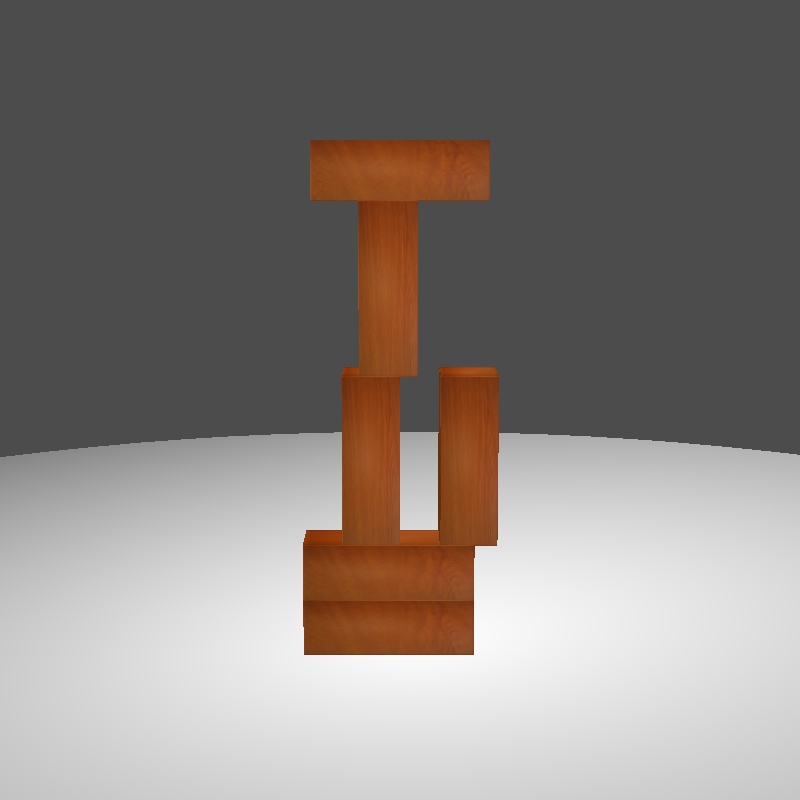}\caption{6 Blocks}\end{subfigure}&
\begin{subfigure}{0.101\textwidth}\centering\includegraphics[width=0.96\columnwidth]{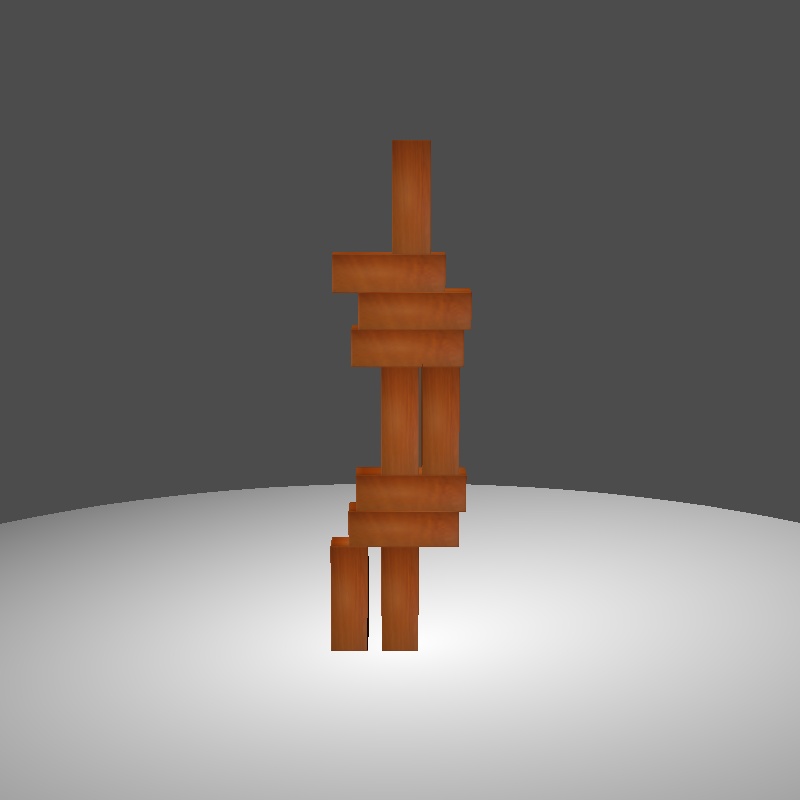}\caption{10 Blocks}\end{subfigure}&
\begin{subfigure}{0.101\textwidth}\centering\includegraphics[width=0.96\columnwidth]{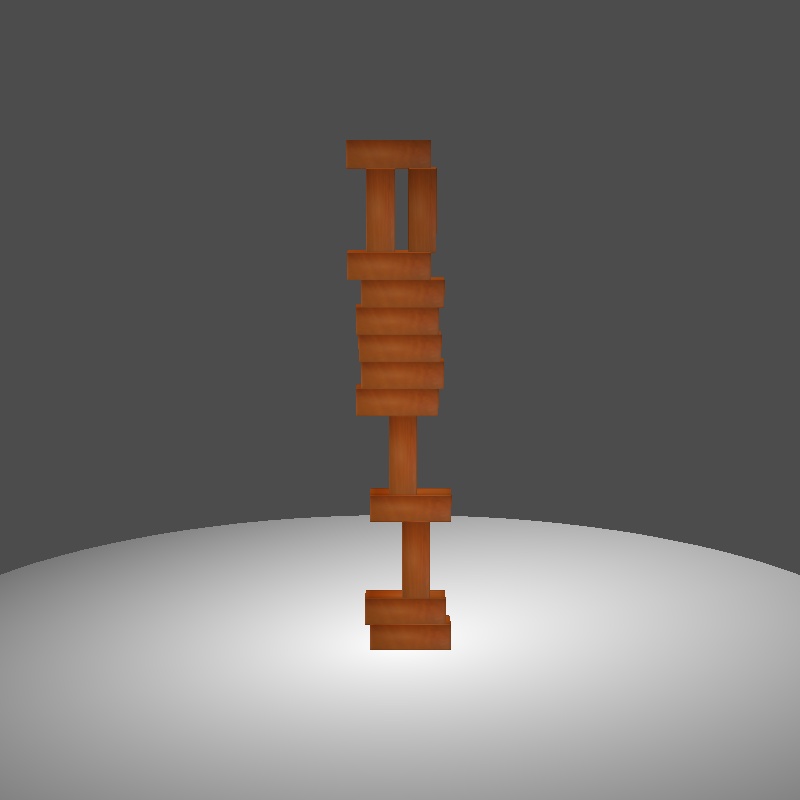}\caption{14 Blocks}\end{subfigure}&
\begin{subfigure}{0.101\textwidth}\centering\includegraphics[width=0.96\columnwidth]{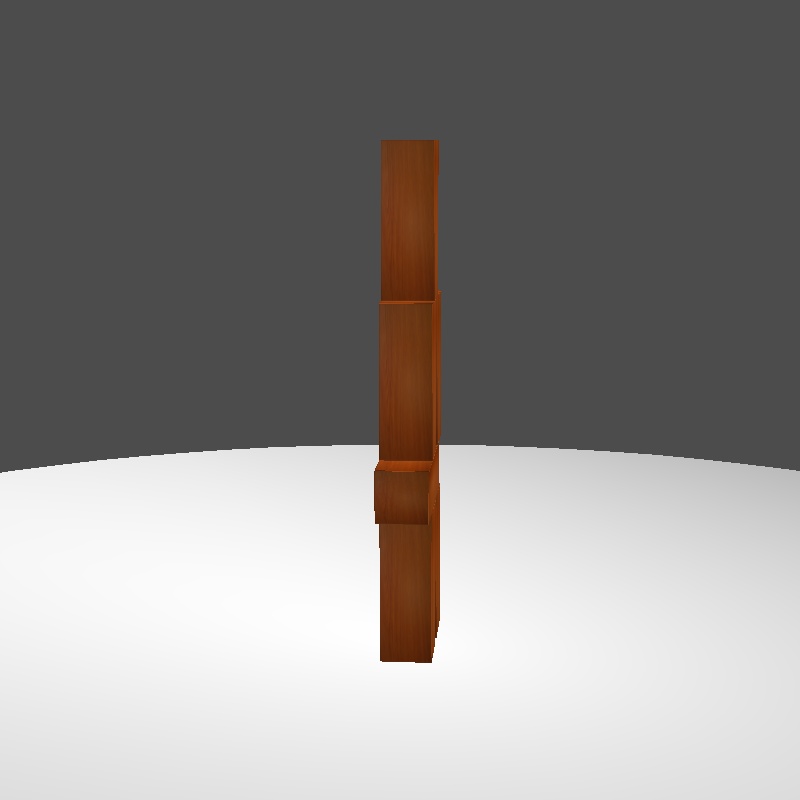}\caption{2D-stack}\end{subfigure}&
\begin{subfigure}{0.101\textwidth}\centering\includegraphics[width=0.96\columnwidth]{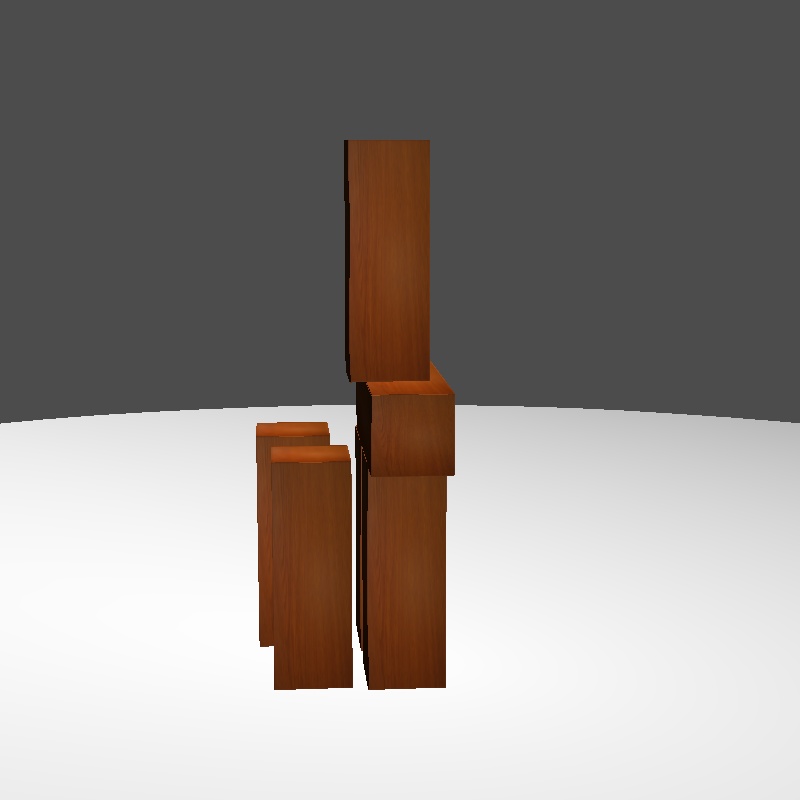}\caption{3D-stack}\end{subfigure}&
\begin{subfigure}{0.101\textwidth}\centering\includegraphics[width=0.96\columnwidth]{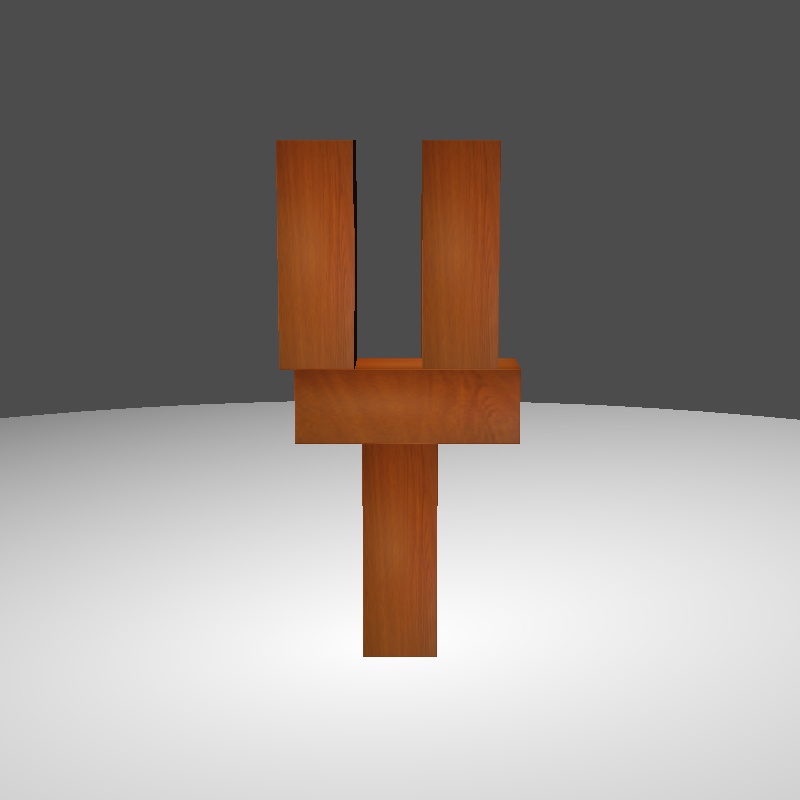}\caption{Size-fix}\end{subfigure} &
\begin{subfigure}{0.101\textwidth}\centering\includegraphics[width=0.96\columnwidth]{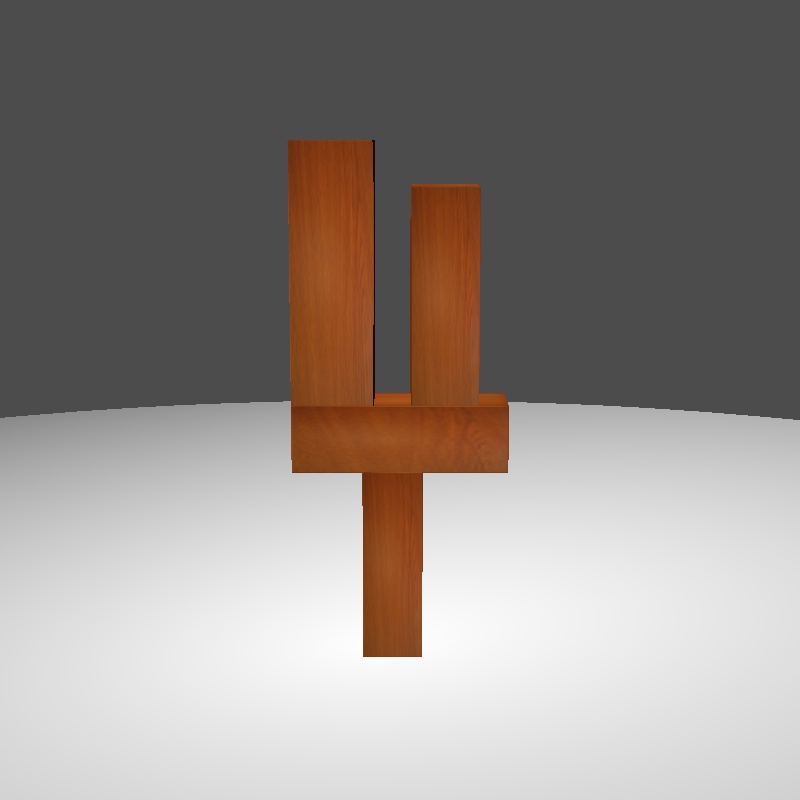}\caption{Size-Vary}\end{subfigure}\\
\hline
\end{tabular}
\caption{Overview of the scene parameters in our rendered scenes. There are 3 groups of scene parameters across number of blocks, stacking depth and block size.}
\label{tab:scene_param}
\end{table*}

\subsection{Stability Prediction from Still Images}
\paragraph{Inspiration from Human Studies}
Research in \cite{hamrick2011internal,battaglia2013simulation} suggests the combinations of the most salient features in the scenes are insufficient to capture people's judgments, however, contemporary study reveals human's perception of visual information, in particular some geometric feature, like critical angle \cite{cholewiak2013visual,cholewiak2015perception} plays an important role in the process. Regardless of the actual inner mechanism for humans to parse the visual input, it is clear there is a mapping $f$ involving visual input $I$ to the stability prediction $P$. 
\begin{gather*}
\begin{aligned}
f: I,\ast \rightarrow P
\end{aligned}
\end{gather*}
Here, $\ast$ denotes other possible information, i.e., the mapping can be inclusive, as in \cite{hamrick2011internal} using it along with other aspects, like physical constraint to make judgment or the mapping is exclusive, as in \cite{cholewiak2013visual} using visual cues alone to decide.

\paragraph{Image Classifier for Stability Prediction}
In our work, we are interested in the mapping $f$ exclusive to visual input and directly predicts the physical stability. To this end, we use deep convolutional neural networks as it has shown great success on image classification tasks \cite{krizhevsky2012imagenet}. Such networks have been shown to be able to adapt to a wide range of classification and prediction task \cite{razavian2014cnn} through re-training or adaptation by fine-tuning.
Therefore, these approaches seem to be adequate methods to study visual prediction on this challenging task with the motivation that by changing conventional image classes labels to stability labels the network can learn ``physical stability salient'' features.

In a pilot study, we tested on a subset of the generated data with LeNet \cite{lecun1995comparison}, a relatively small network designed for digit recognition, AlexNet \cite{krizhevsky2012imagenet}, a large network and VGG Net\cite{simonyan2014very}, an even larger network than AlexNet. We trained from scratch for the LeNet and fine-tuned for the large network pre-trained on ImageNet \cite{deng2009imagenet}. VGG Net consistently outperforms the other two, hence we use it across our experiment. We use the Caffe framework \cite{jia2014caffe} in all our experiments. 

\subsection{Prediction Performance}
In this part of the experiments, the images are captured before the physics engine is enabled, and the stability labels are recorded from the simulation engine as described before. At the training time, the model has access to the images and the stability labels. At test time, the learned model predicts the stability results against the results generated by the simulator.

We divide the experiment design into 3 sets: the intra-group, cross-group and generalization. The first set investigates influence on the model's performance from an individual scene parameter, the other two sets explore generalization properties under different settings.

%--------------------------------------------------------------------------------------------------------------------------------------
\subsubsection{Intra-Group Experiment}
In this set of experiments, we train and test on the scenes with the same scene parameters in order to assess the feasibility of our task.

\paragraph{Number of Blocks (4B, 6B, 10B, 14B)}
In this group of experiment, we fix the stacking depth and keep the all blocks in the same size but vary the number of blocks in the scene to observe how it affects the prediction rates from the image trained model, which approximates the relative recognition difficulty from this scene parameter alone. The results are shown in Table~\ref{tab:intra_group}. A consistent drop of performance can be observed with increasing number of blocks in the scene under various block sizes and stacking depth conditions. More blocks in the scene generally leads to higher scene structure and hence higher difficulty in perception. 

\paragraph{Block Size (Uni. vs. NonUni.)}
In this group of experiment, we aim to explore how same size and varied blocks sizes affect the prediction rates from the image trained model. We compare the results at different number of blocks to the previous group, in the most obvious case,  scenes happened to have similar stacking patterns and same number of blocks can result in changes visual appearance. To further eliminate the influence from the stacking depth, we fix all the scenes in this group to be 2D stacking only. As can be seen from Table~\ref{tab:intra_group}, the performance decreases when moving from 2D stacking to 3D. The additional variety introduced by the block size indeed makes the task more challenging.

\paragraph{Stacking Depth (2D vs. 3D)}
In this group of experiment, we investigate how stacking depth affects the prediction rates. With increasing stacking depth, it naturally introduces ambiguity in the perception of the scene structure, namely some parts of the scene can be occluded or partially occluded by other parts. Similar to the experiments in previous groups, we want to minimize the influences from other scene parameters, we fix the block size to be the same and only observe the performance across different number of blocks. The results in Table~\ref{tab:intra_group} show a little inconsistent behaviors between relative simple scenes (4 blocks and 6 blocks) and difficult scenes (10 blocks and 14 blocks). For simple scenes, prediction accuracy increases when moving from $2D$ stacking to $3D$ while it is the other way around for the complex scene. Naturally relaxing the constraint in stacking depth can introduce additional challenge for perception of depth information, yet given a fixed number of blocks in the scene, the condition change is also more likely to make the scene structure lower which reduces the difficulty in perception. A combination of these two factors decides the final difficulty of the task, for simple scenes, the height factor has stronger influence and hence exhibits better prediction accuracy for $3D$ over $2D$ stacking while for complex scenes, the stacking depth dominates the influence as the significant higher number of blocks can retain a reasonable height of the structure, hence receives decreased performance when moving from $2D$ stacking to $3D$.

\begin{table}\setlength{\tabcolsep}{7pt}
\centering
\ra{1.4}
\begin{tabular*}{0.8\linewidth}{@{ }ccccc@{ }}\toprule
Num.of Blks & \multicolumn{2}{c}{Uni.} & \phantom{ab}& \multicolumn{1}{c}{NonUni.}\\
\cmidrule{2-3} \cmidrule{5-5}
 & $2D$ & $3D$ && $2D$\\ \midrule
$4B$ & 93.0 & 99.2 && 93.2\\
$6B$ & 88.8& 91.6&& 88.0\\
$10B$ & 76.4& 68.4&& 69.8\\
$14B$ & 71.2& 57.0&& 74.8\\ \bottomrule
\end{tabular*}
\caption{Intra-group experiment by varying scene parameters.} 
\label{tab:intra_group}
\end{table}

%--------------------------------------------------------------------------------------------------------------------------------------
\subsubsection{Cross-Group Experiment}
In this set of experiment, we want to see how the learned model transfers across scenes with different complexity, so we further divide the scene groups into two large groups by the number of blocks, where a {\it simple scene} group for all the scenes with $4$ and $6$ blocks and a {\it complex scene} for the rest of scenes with $10$ and $14$ blocks. We investigate in two-direction classification, shown in the figure in Table~\ref{tab:cross_group}:
\begin{enumerate}
\item
Train on simple scenes and predict on complex scenes: Train on 4 and 6 blocks and test on 10 and 14 blocks
\item
Train on complex scenes and predict on simple scenes: Train on 10 and 14 blocks and test on 4 and 6 blocks
\end{enumerate}

As shown in Table~\ref{tab:cross_group}, when trained on simple scenes and predicting on complex scenes, it gets $69.9\%$, which is significantly better than random guess at $50\%$. This is understandable as the learned visual feature can transfer across different scene. Further we observe significant performance boost when trained on complex scenes and tested on simple scene. This can be explained by the richer feature learned from the complex scenes with better generalization. 

\begin{table}\setlength{\tabcolsep}{7pt}
\centering
\ra{1.4}
\begin{tabular*}{0.9\linewidth}{l c c}
\multicolumn{3}{c}{\includegraphics[width=0.8\linewidth]{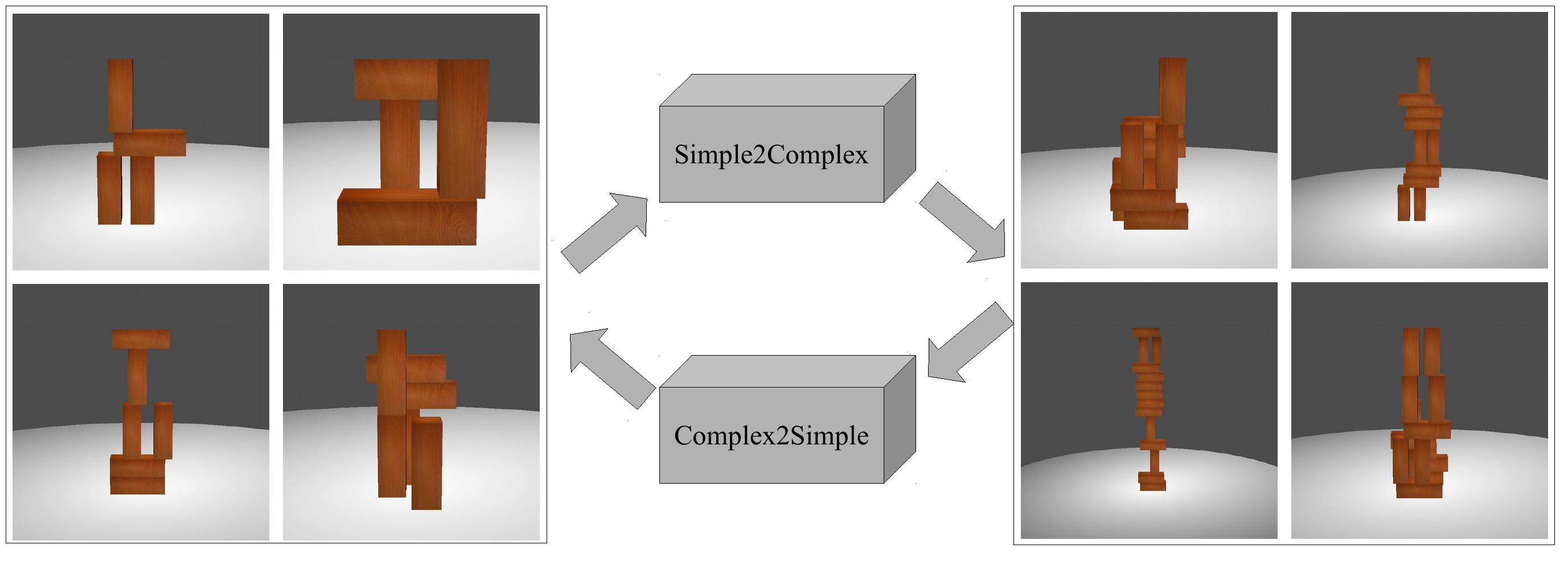}}\\
\toprule
Setting & Simple $\rightarrow$ Complex & Complex $\rightarrow$ Simple\\
\hline
Accuracy (\%)& 69.9 & 86.9\\
\bottomrule
\end{tabular*}
\caption{The upper figure shows the experiment settings for Cross-group classification where we train on simpler scenes and test on more complex scenes. The lower table shows the results.}
\label{tab:cross_group}
\end{table}

%--------------------------------------------------------------------------------------------------------------------------------------
\subsubsection{Generalization Experiment}
In this set of experiment, we want to explore if we can train a general model to predict stability for scenes with any scene parameters, which is very similar to human's prediction in the task. We use training images from all different scene groups and test on any groups. The Result is shown in Table~\ref{tab:general}. While the performance exhibits similar trend to the one in the intra-group with respect to the complexity of the scenes, namely increasing recognition rate for simpler settings and decreasing rate for more complex settings, there is a consistent improvement over the intra-group experiment for individual groups. Together with the result in the cross-group experiment, it suggests  a strong generalization capability of the image trained model.

\begin{table}\setlength{\tabcolsep}{7pt}
\centering
\ra{1.4}
\begin{tabular*}{0.85\linewidth}{@{ }cccccc@{ }}\toprule
Num.of Blks & \multicolumn{2}{c}{Uni.} & \phantom{ab}& \multicolumn{2}{c}{NonUni.}\\
\cmidrule{2-3} \cmidrule{5-6}
 & $2D$ & $3D$ && $2D$ & $3D$\\ 
 \midrule
4B  &93.2 &99.0 &&95.4 &99.8 \\
6B  &89.0 &94.8 &&87.8 &93.0 \\
10B &83.4 &76.0 &&77.2 &74.8 \\
14B &82.4 &67.2 &&78.4 &66.2 \\ 
\bottomrule
\end{tabular*}
\caption{Results for generalization experiments.} 
\label{tab:general}
\end{table}

\begin{table}\setlength{\tabcolsep}{7pt}
\centering
\ra{1.3}
\begin{tabular*}{1.\linewidth}{@{ }cccccc@{ }}\toprule
Num.of Blks & \multicolumn{2}{c}{Uni.} & \phantom{a}& \multicolumn{2}{c}{NonUni.}\\
\cmidrule{2-3} \cmidrule{5-6}
& $2D$ & $3D$ && $2D$ & $3D$\\ 
 \midrule
4B  &79.1/{\bf91.7} &93.8/{\bf 100.0} &&72.9/{\bf 93.8} &92.7/{\bf 100.0} \\
6B  &78.1/{\bf 91.7} &83.3/{\bf 93.8} &&71.9/{\bf 87.5} &89.6/{\bf 93.8} \\
10B &67.7/{\bf 87.5} &72.9/72.9 &&66.7/{\bf 72.9} &71.9/68.8 \\
14B &71.9/{\bf 79.2} &68.8/66.7 &&71.9/{\bf 81.3} &59.3/{\bf 60.4} \\ 
\bottomrule
\end{tabular*}
\caption{Results from human subject test $a$ and corresponded accuracies from image-based model $b$ in format $a/b$ for the sampled data.} 
\label{tab:human_test}
\end{table}

\subsubsection{Discussion}
Overall, we can conclude that direct stability prediction is possible and in fact fairly accurate at recognition rates over $80\%$ for moderate difficulty levels. As expected, the 3D setting adds difficulties to the prediction from appearance due to significant occlusion for towers of more than 10 blocks. Surprisingly, little effect was observed for small tower sizes switching from uniform to non-uniform blocks - although the appearance difference can be quite small.
To better understand our results, we further discuss the following two questions:

\begin{figure*}
\centering
	\begin{subfigure}[b]{0.6\linewidth}
		\centering
		\includegraphics[width=1.\linewidth]{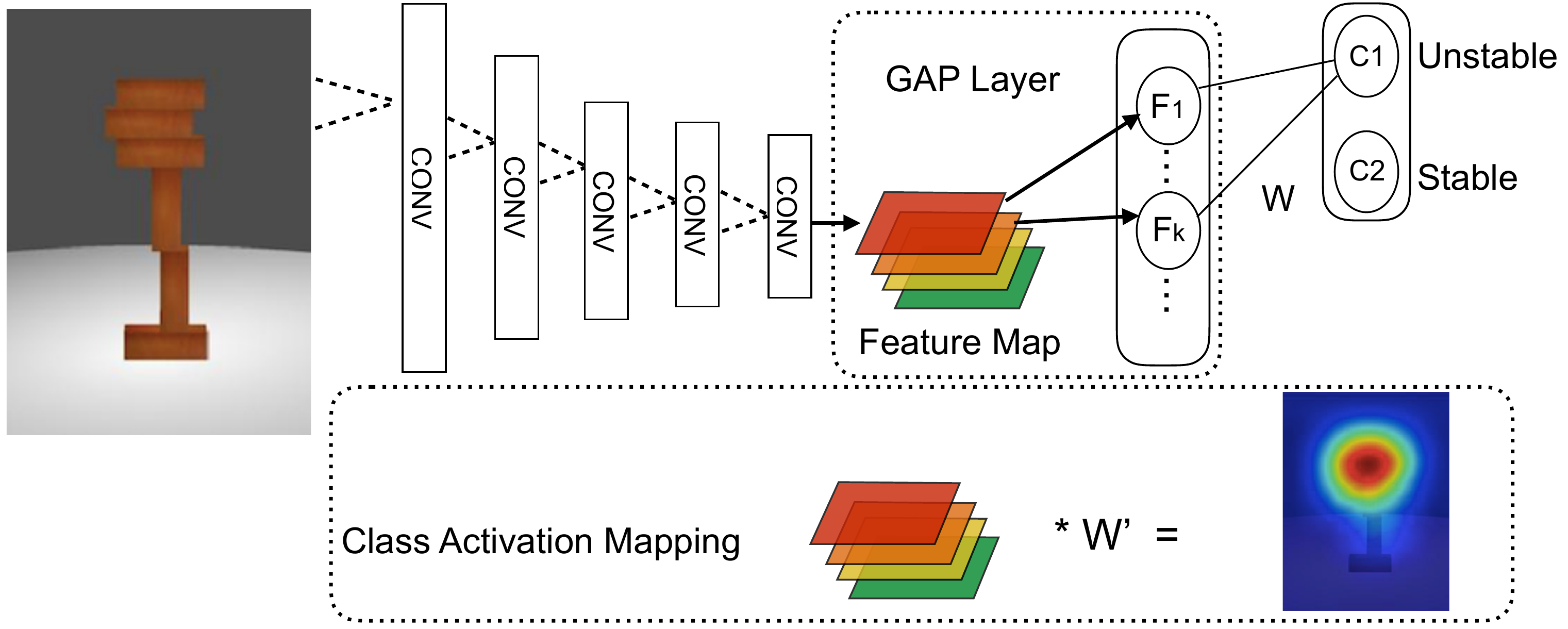}
		\caption{By introducing the GAP layer directly connected to the final output, the learned weights can be backprojected to the feature map for each category to construct the CAM. The CAM can be used to visualize the discriminative image regions for individual category.}
	\label{fig:gap_pipe}
	\end{subfigure}
	\begin{subfigure}[b]{0.34\linewidth}
		\centering
		\includegraphics[width=1.\linewidth]{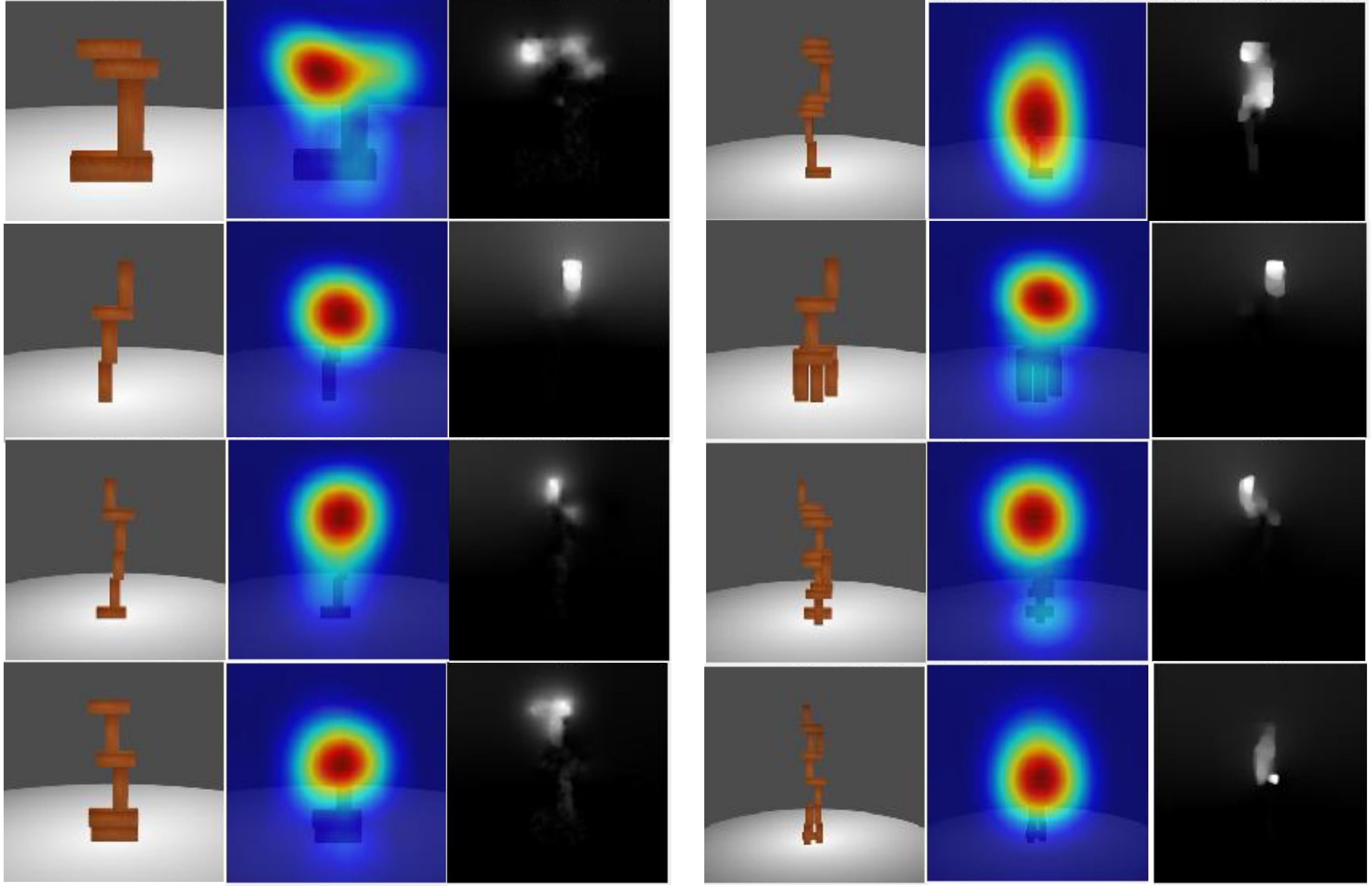}
		\caption{Examples of CAM showing the discriminative regions for unstable prediction in comparison to the flow magnitude indicating where the collapse motion begins.  For each example, from left to right are original image, CAM and flow magnitude map. }
	\label{fig:gap_example}
	\end{subfigure}
\caption{We use CAM to visualize the results for model interpretation.}
\label{fig:gap}
\end{figure*}

\textbf{How does the model performs compared to human?} To answer this, we conduct a human subject test. We recruit human subjects to predict stability for give scene images. Due to large number of test data, we sample images from different scene groups for human subject test. 8 subjects are recruited for the test. Each subject is presented with a set of captured images from the test split. Each set includes $96$ images where images cover all $16$ scene groups with $6$ scene instances per group. For each scene image, subject is required to rate the stability on a scale from $1-5$ without any constraint for response time:
\begin{enumerate}
\item
Definitely unstable: definitely at least one block will move/fall
\item
Probably unstable: probably at least one block will move/fall
\item
Cannot tell: the subject is not sure about the stability
\item
Probably stable: probably no block will move/fall
\item
Definitely stable: definitely no block will move/fall
\end{enumerate}

The predictions are binarized, namely 1) and 2) are treated as unstable prediction, 4) and 5) as stable prediction, ``Cannot tell'' will be counted as $0.5$ correct prediction. 

The results are shown in Table~\ref{tab:human_test}. For simple scenes with few blocks, human can reach close to perfect performance while for complex scenes, the performance drops significantly to around $60\%$. Compared to  human prediction in the same test data, the image-based model outperforms human in most scene groups. While showing similar trends in performance with respect to different scene parameters, the image-based model is less affected by a more difficult scene parameter setting, for example, given the same block size and stacking depth condition, the prediction accuracy decreases more slowly than the counter part in human prediction. We interpret this as image-based model possesses better generalization capability than human in the very task. 

\textbf{ Does the model learn something explicitly interpretable?}  Here we apply the technique from \cite{zhou2015cnnlocalization} to visualize the learned discriminative image regions from CNN for individual category. The approach is illustrated in Figure~\ref{fig:gap_pipe}. With Global Average Pooling (GAP), the resulted spatial average of the feature maps from previous convolutional layers forms fully-connected layer to directly decides the final output. By back-projecting the weights from the fully-connected layer from each category, we can hence obtain Class Activation Map (CAM) to visualize the discriminative image regions. In our case, we investigate discriminative regions for unstable predictions to see if the model can spot the weakness in the structure. We use deep flow\cite{weinzaepfel2013deepflow} to compute the optical flow magnitude between the frame before the physics engine is enabled and the one afterwards to serve as a coarse ground truth for the structural weakness where we assume the collapse motion starts from such weakness in the structure. Though not universal among the unstable cases, we do find significant positive cases showing high correlation between the activation regions in CAM for unstable output and the regions where the collapse motion begins. Some examples are shown in Figure~\ref{fig:gap_example}.
\section{From Visual Stability Test to Manipulation}
\begin{figure}
\centering
\includegraphics[width=0.9\linewidth]{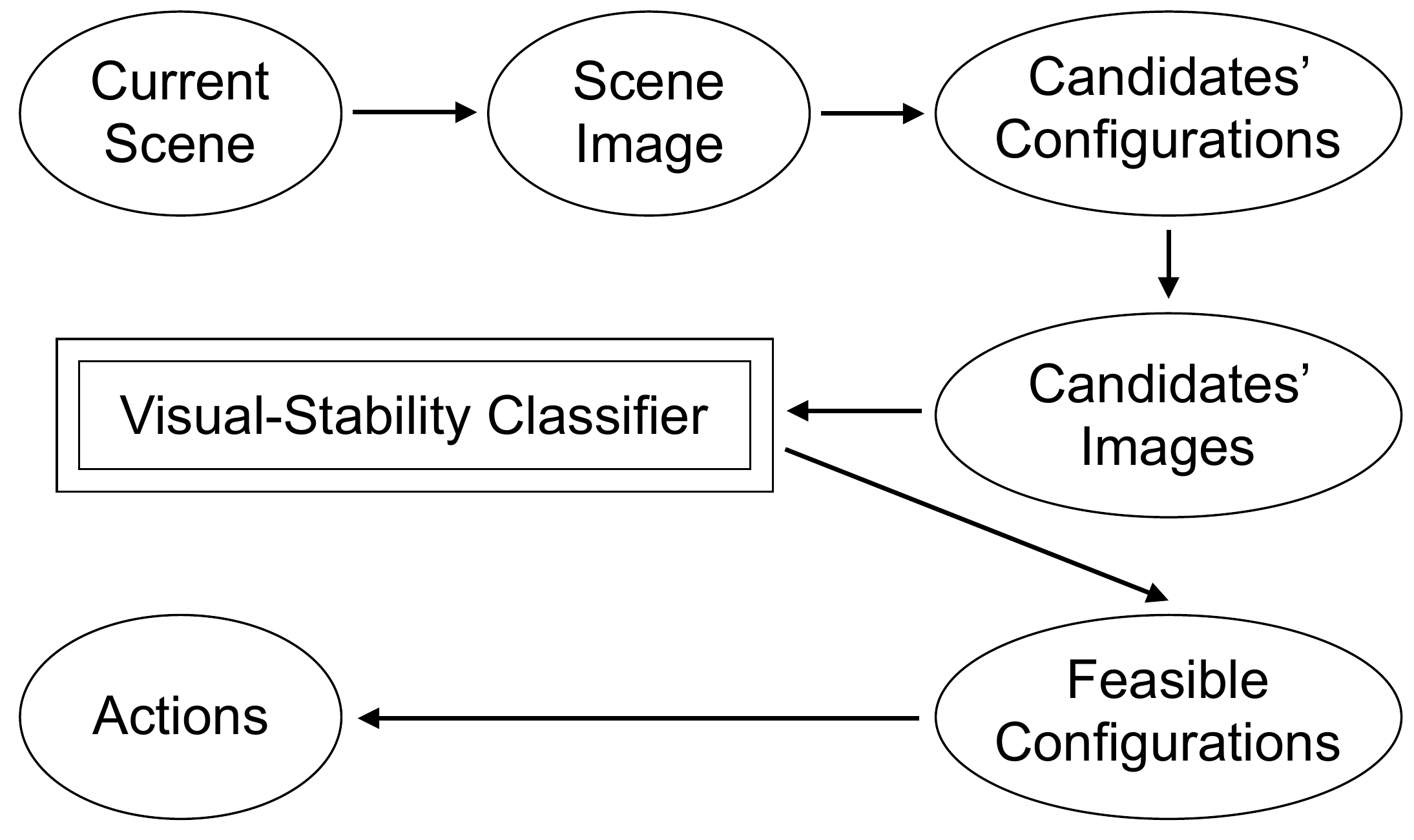}
\caption{An overview of our manipulation system.}
\label{fig:manip}
\end{figure}

\begin{figure}
\centering
\begin{subfigure}[b]{0.61\linewidth}\includegraphics[width=1\linewidth]{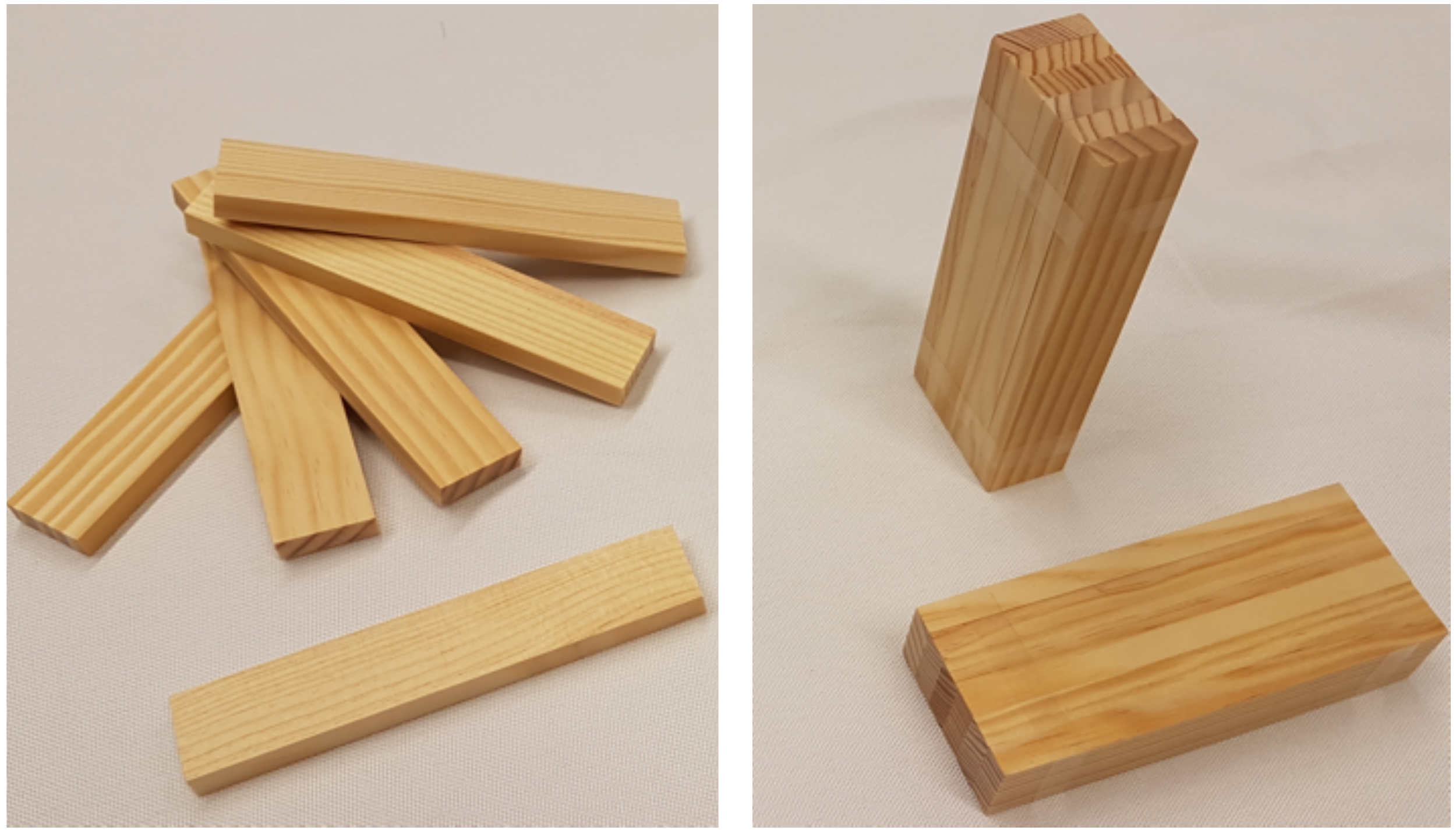}
\caption{Kapla block (left), block in test (right).}
\label{fig:blks}\end{subfigure}
\begin{subfigure}[b]{0.34\linewidth}\includegraphics[width=1\linewidth]{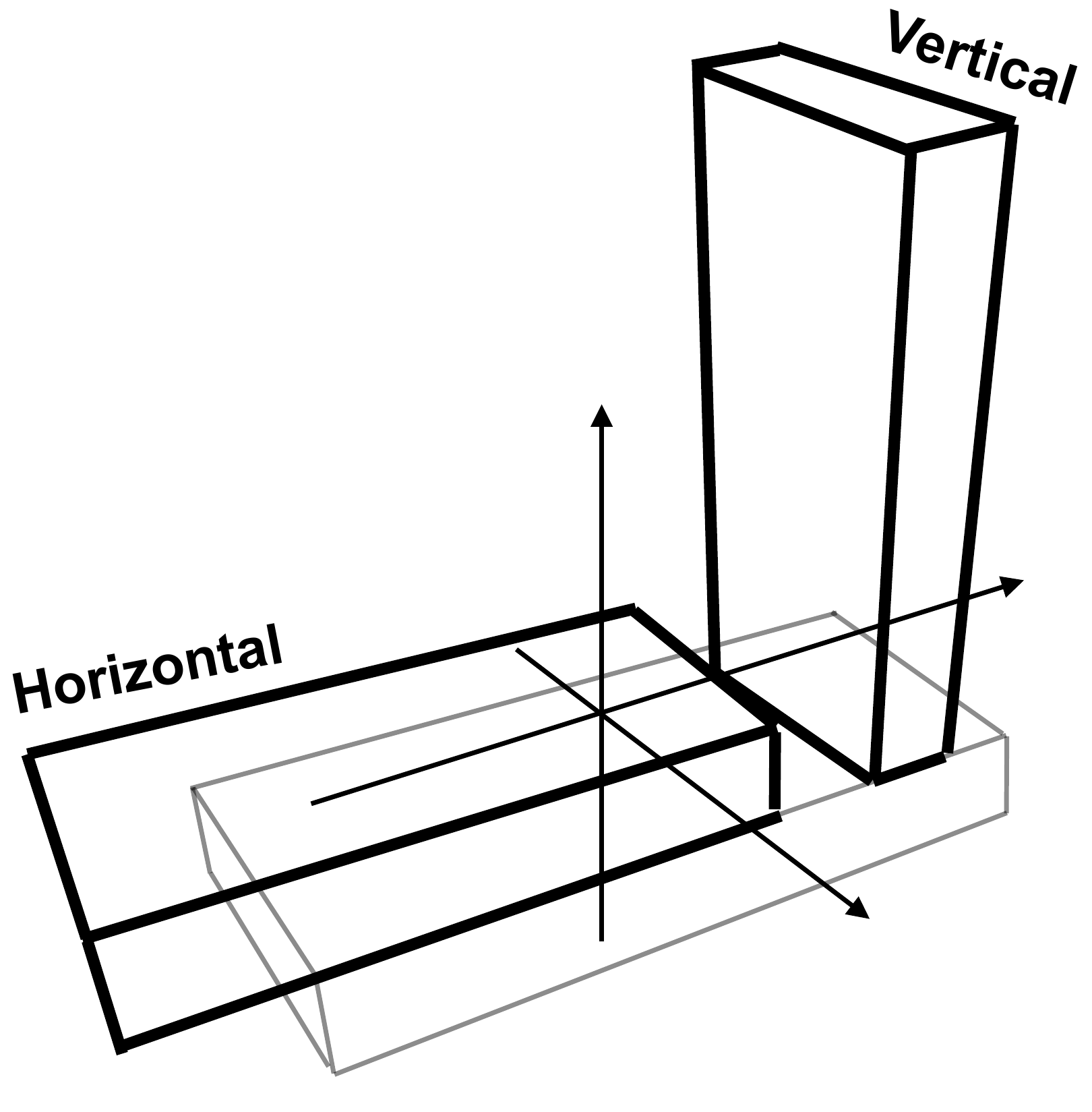}
\caption{Allowed configurations in test.}
\label{fig:blk_config}\end{subfigure}
\caption{Blocks used in our experiment.}
\label{fig:overview_manip}
\end{figure}

In previous section, we have shown that an appereance-based model can predict  physical stability relatively well on the synthetic data. Now we want to further explore if and how the synthetic data trained model can be utilized for a real world application, especially for robotic manipulation. 
Hence, we decide to set up a testbed where a Baxter robot's task is to stack one wood block on a given block structure without breaking the structure's stability as shown in Figure~\ref{fig:teaser}. The overview of our system is illustrated in Figure~\ref{fig:manip}. In our experiment, we use Kapla blocks as basic unit, and tape 6 blocks into a bigger one as shown in Figure~\ref{fig:blks}. To simplify the task, adjustments were made to the free-style stacking:

\begin{itemize}
\item The given block structure is restricted to be single layer as the ${\text 2D}$ case in the previous section. For the final test, we report results on the 6 scenes as shown in Table~\ref{tab:real_task}.
\item The block to be put on top of the given structure is limited two canonical configurations \{vertical, horizontal\} as shown in Figure~\ref{fig:blk_config}. and assumed to be held in hand of robot before the placement.
\item The block is constrained to be placed on the top-most horizontal surface (stacking surface) in the given structure.
\item The depth of the structure (perpendicular distance to the robot) is calibrated so that we only need to decide the horizontal and vertical displacements with respect to the stacking surface. 
\end{itemize}

\subsection{Prediction on Real World Data}
Considering there are significant difference between the synthesized data and real world captured data, including factors (not limited to) like the texture, the illumination condition, size of blocks and accuracy of the render, we performed a pilot study to directly apply the model trained on the RGB images to predict stability on the real data, but only got results on par with random guessing. Hence we decided to train the visual-stability model on the binary-valued foreground mask on the synthesized data and deal with the masks at test time also for the real scenes. In this way, we significantly reduce the effect from the aforementioned factors. Observing comparable results when using the RGB images, we continue to the approach on real world data.

\begin{figure}
\centering
\includegraphics[width=1\linewidth]{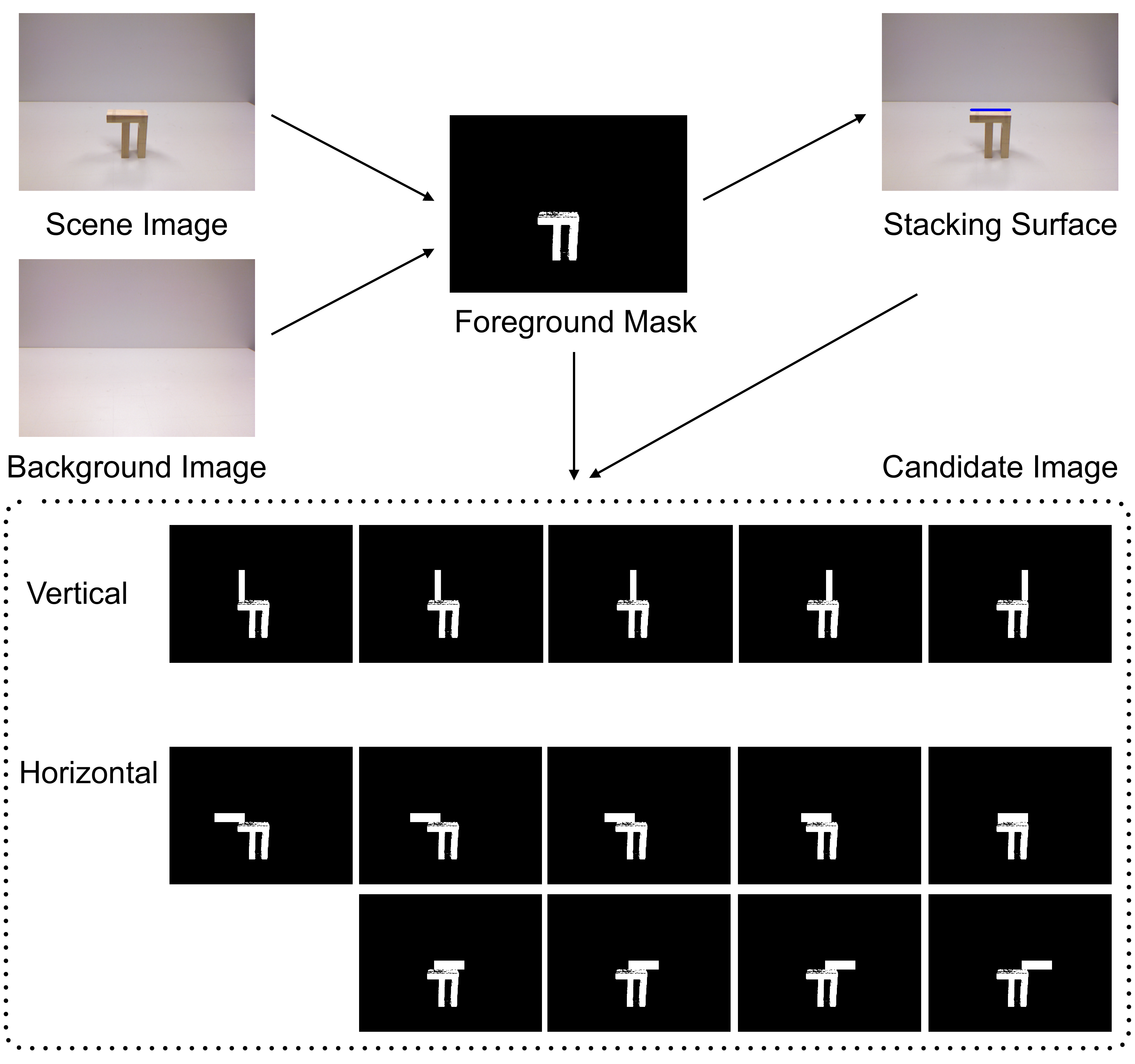}
\caption{The procedure to generate candidates placement images for a give scene in our experiment.}
\label{fig:candidates}
\end{figure}

At test time, a background image is first captured for the empty scene. Then for each test scene as shown in Table~\ref{tab:real_task}, an image is captured and converted to foreground mask via background subtraction. The top-most horizontal boundary is detected as the stacking surface and then used to generate candidate placements: the surface is divided evenly into 9 horizontal candidates and 5 vertical candidates, so overall there are 84 candidates. The process is shown in Figure~\ref{fig:candidates}. Afterwards, these candidates are put to the visual-stability model for stability prediction. Each generated candidate's actual stability is manually tested and recorded as ground truth. The final recognition result is shown in Table~\ref{tab:real_task}. The model trained with synthetic data is able to predict with overall accuracy of $78.6\%$ across different candidates in real world.

\subsection{Manipulation Test}
At test time, when the model predicts a give candidate placement as stable, the robot will execute routine to place the block with 3 attempts. We count the execution as a success if any of the attempt works. The manipulation success rate is defined as:
\begin{gather*}
\begin{aligned}
\frac{\text{\#\{successful placements\}}}{\text{\#\{all stable placements\}}}
\end{aligned}
\end{gather*}
where $\text{\#\{successful placements\}}$ is the number of successful placements made by the robot, and $\text{\#\{all stable placements\}}$ is the number of all ground truth stable placements.

As shown in Table~\ref{tab:real_task}, the manipulation performance is good across most of the scenes for both horizontal and vertical placements except for the 6-th scene where the classifier predicts all candidates as unstable hence no attempts have been made by the robot.

\subsection{Discussion}
Comparing to the work in block manipulation \cite{wang2009robot}, we do not fit 3D models or run physics simulation at test time for the given scene but instead use the scene image as input to directly predict the physics of the structure. Simply putting the block along the center of mass (COM) of the given structure may often be a feasible option, yet, there are two limitations to this approach: first, it is nontrivial to compute the COM of a given structure; second, it only gives one possible stable solution (assuming it actually stay stable). In comparison, our method does not rely the COM of the structure and provide a search over multiple possible solutions. %As a future work, we can extend our method to sequentially stacking multiple blocks.

\begin{table*}\setlength{\tabcolsep}{4pt}
\centering
\ra{1.4}
\begin{tabular*}{1\linewidth}{crrrrrrrrrrrr}\toprule
Id. & \multicolumn{2}{c}{1} & \multicolumn{2}{c}{2} & \multicolumn{2}{c}{3} & \multicolumn{2}{c}{4} &\multicolumn{2}{c}{5} &\multicolumn{2}{c}{6}\\
\hline 
&&&&&&\\
Scene & 
\multicolumn{2}{c}{\includegraphics[width=0.13\linewidth]{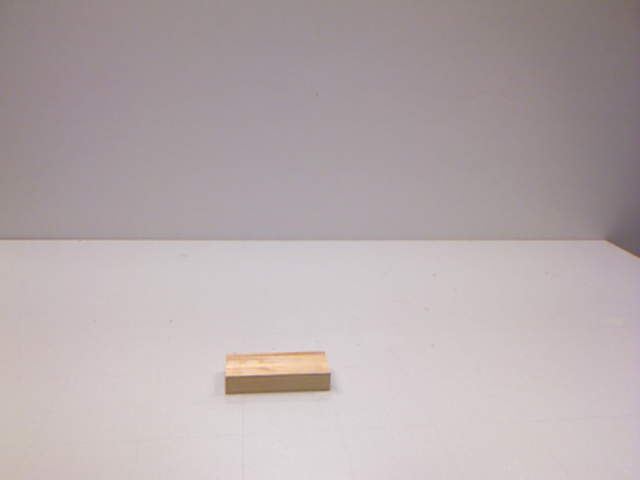}} &
\multicolumn{2}{c}{\includegraphics[width=0.13\linewidth]{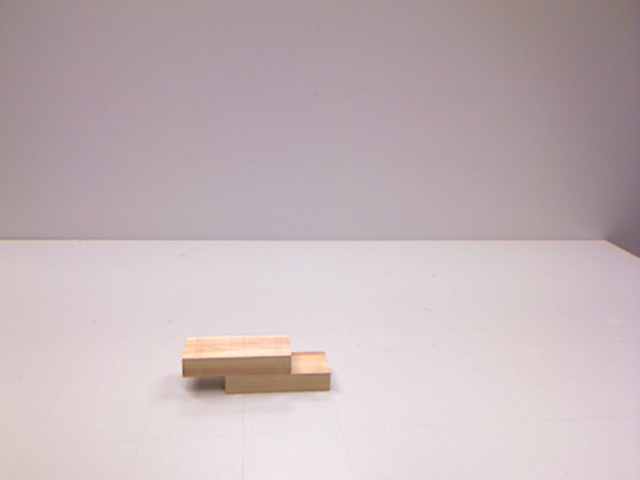}} &
\multicolumn{2}{c}{\includegraphics[width=0.13\linewidth]{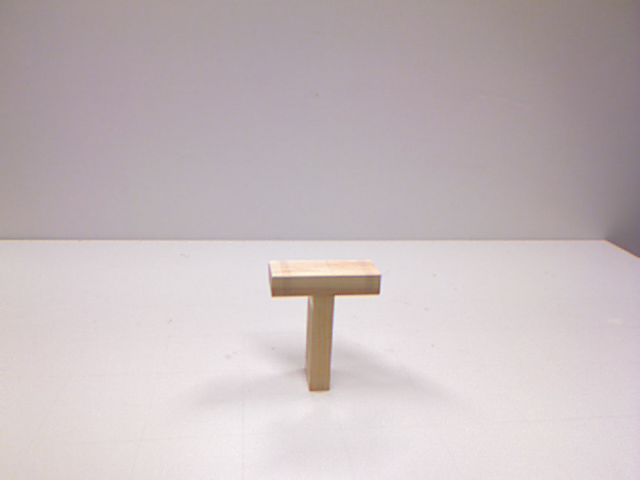}} &
\multicolumn{2}{c}{\includegraphics[width=0.13\linewidth]{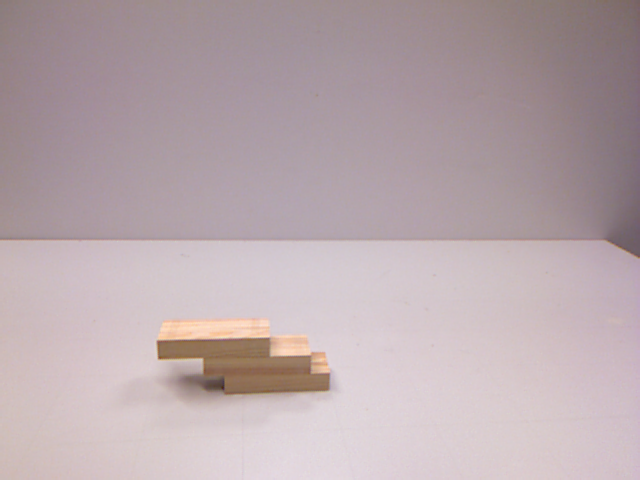}} &
\multicolumn{2}{c}{\includegraphics[width=0.13\linewidth]{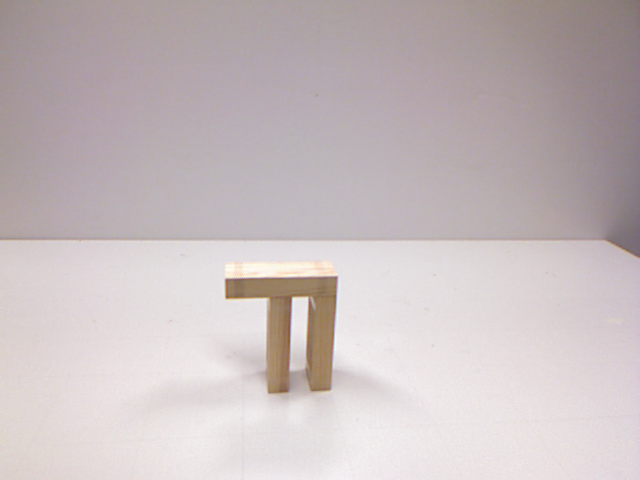}} &
\multicolumn{2}{c}{\includegraphics[width=0.13\linewidth]{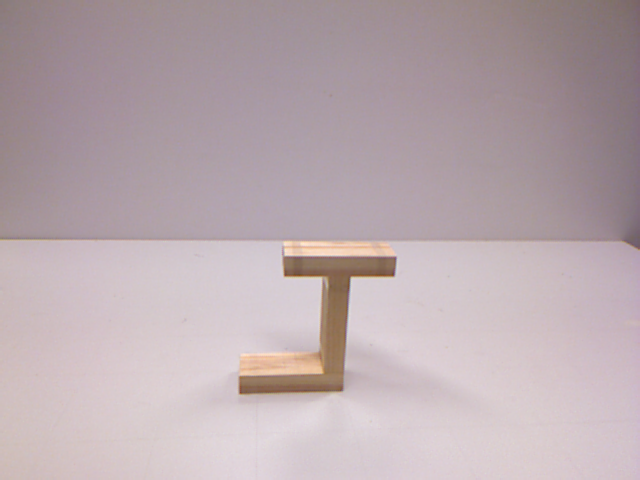}} \\
\hline
Pred.($\%$)& 
66.7& 100.0& 
66.7& 60.0& 
88.9& 100.0& 
77.8& 80.0& 
100.0& 40.0& 
66.7& 60.0\\
Mani.($\%$)&
80.0(4/5)& 100.0(5/5)&
66.7(2/3)& 100.0(3/3)&
66.7(2/3)& 100.0(1/1)&
66.7(2/2)& 66.7(2/3)&
100.0(3/3)& 25.0(1/4)&
0.0(0/3)& 0.0(0/1)\\
\bottomrule
Placement& 
H& V&
H& V&
H& V&
H& V&
H& V&
H& V\\
\hline
\end{tabular*}
\caption{Results for real world test. ``Pred.'' is the prediction accuracy. ``Mani.'' is the manipulation success rate with counts for successful placements/all possible stable placements for each scene. ``H/V'' refer to horizontal/vertical placement.} % from prediction files
\label{tab:real_task}
\end{table*}
\section{Conclusion}
In this work, we answer the question if and how well we can build up a mechanism to predict physical stability directly from visual input. In contrast to existing approaches, we bypass explicit 3D representations and physical simulation and learn a model for visual stability prediction from data. We evaluate our model on a range of conditions including variations in number of blocks, size of blocks and 3D structure of the overall tower. The results reflect the challenges of inference with growing complexity of the structure. To further understand the results, we conduct a human subject study on a subset of our synthetic data and show that our model achieves comparable or even better results than humans in the same setting. Moreover, we investigate the discriminative image regions found by the model and spot correlation between such regions and initial collapse area in the structure.
Finally, We apply our approach to a block stacking setting and show that our model can guide a robot for placements of new blocks by  predicting the stability of future states.

%%%%%%%%%%%%%%%%%%%%%%%%%%%%%%%%%%%%%%%%%%%%%%%%%%%%%%%%%%%%%%%%%%%%%%%%%%%%%%%%

\bibliographystyle{IEEEtran} % use IEEEtran.bst style
\bibliography{mybib}

\end{document}